\documentclass[sigconf,screen]{acmart}

\AtBeginDocument{%
  \providecommand\BibTeX{{%
    \normalfont B\kern-0.5em{\scshape i\kern-0.25em b}\kern-0.8em\TeX}}}
\setcopyright{acmlicensed}
\copyrightyear{2025}
\acmYear{2025}
\acmDOI{XXXXXXX.XXXXXXX}
\renewcommand\footnotetextcopyrightpermission[1]{} 
\usepackage[normalem]{ulem}
\useunder{\uline}{\ul}{}

\usepackage{bm}
\usepackage{multirow}
\usepackage{tabularx}
\usepackage{ragged2e}
\usepackage{amsmath,amsfonts, bm,mathtools}

\usepackage{graphicx}

\usepackage{float}
\usepackage[ruled,linesnumbered]{algorithm2e}
\usepackage{algorithmicx}
\usepackage{algpseudocode}
\usepackage{array}
\usepackage{wasysym}
\usepackage{textcomp}
\usepackage{stfloats}

\usepackage{url}
\usepackage{verbatim}
\DeclareMathOperator*{\argmin}{arg\,min}

\usepackage{booktabs}
\usepackage{multirow}
\usepackage{color, colortbl}
\usepackage{subcaption}
\usepackage{tabu}
\usepackage{pifont}
\usepackage{makecell}
\usepackage{paralist}
\usepackage{threeparttable}
\usepackage{enumitem}
\usepackage{hhline}
\usepackage{scalerel,xparse}

\acmConference[MM '25]{Proceedings of the 33rd ACM International Conference on Multimedia}{October 27--October 31, 2025}{Dublin, Ireland}
\acmBooktitle{Proceedings of the 33rd ACM International Conference on Multimedia(MM '25),
 October 27--October 31, 2025, Dublin, Ireland} 
\acmISBN{978-1-4503-XXXX-X/25/10}

\begin{document}

\title[DreamFrame]{DreamFrame: Enhancing Video Understanding via Automatically Generated QA and Style-Consistent Keyframes}

\author{Zhende Song}

\email{zdsong23@m.fudan.edu.cn}
\affiliation{
  \institution{Fudan University}
  \city{Shanghai}
  \country{China}
}

\author{Chenchen Wang}

\email{24210720042@m.fudan.edu.cn}
\affiliation{
  \institution{Fudan University}
  \city{Shanghai}
  \country{China}
}

\author{Jiamu Sheng}

\email{jmsheng22@m.fudan.edu.cn}
\affiliation{
  \institution{Fudan University}
  \city{Shanghai}
  \country{China}
}

\author{Chi Zhang}

\email{chizhang@westlake.edu.cn}
\authornote{Work done while the author was a researcher at Tencent PCG}
\affiliation{
  \institution{Westlake University, Tencent PCG}
  \city{Hangzhou}
  \country{China}
}

\author{Shengji Tang}
\email{21210720037@m.fudan.edu.cn}
\affiliation{
  \institution{Fudan University}
  \city{Shanghai}
  \country{China}
}

\author{Jiayuan Fan}
\authornote{Corresponding Author.}
\email{jyfan@fudan.edu.cn}
\affiliation{
  \institution{Fudan University}
  \city{Shanghai}
  \country{China}
}

\author{Tao Chen}
\email{eetchen@fudan.edu.cn}
\affiliation{
  \institution{Fudan University}
  \city{Shanghai}
  \country{China}
}

\renewcommand{\shortauthors}{Song, et al.}

\begin{abstract}
Recent large vision-language models (LVLMs) for video understanding are primarily fine-tuned with various videos scraped from online platforms. Existing datasets, such as ActivityNet, require considerable human labor for structuring and annotation before effectively utilized for tuning LVLMs. While current LVLMs are primarily trained on existing datasets in broad, general-purpose settings, adapting them to specific downstream scenarios remains challenging, as collecting and annotating task-specific videos is highly labor-intensive and time-consuming.
To address this issue, we propose a three-stage framework named \textbf{DreamFrame} for automatically generating style-consistent keyframes and corresponding question-answer (QA) pairs to support LVLM instruction tuning. DreamFrame generates datasets in a movie-like manner. First, we utilize an LLM to generate structured movie plots including movie prior information (like overview and style), frame descriptions and plot-related QA pairs, with a story expansion strategy to mitigate context length limitations.Then, to ensure visual consistency across generated frames, we design a Style Immobilization Process which maintains consistent style through an embedding learning strategy. Finally, frame descriptions and style embeddings are integrated to produce coherent keyframes. Using DreamFrame, we construct a dataset comprising approximately 1k stylized keyframe-like videos and 100k diverse QA pairs. Extensive fine-tuned experiments on various LVLM architectures demonstrate the effectiveness of the proposed dataset. Furthermore, based on the proposed dataset, we fine-tune a new LVLM named DreamFrame-7B, which significantly surpasses the previous similar-sized LVLMs (\textbf{+2.2} compared with VideoLLaVA-7B on MvBench) across different benchmarks. Code, data and supplementary will be at \url{https://deaddawn.github.io/DreamFrame}.
\end{abstract}



\maketitle

\section{Introduction}

Recent advancements in multi-modal learning, driven by large language models (LLMs)~\cite{gpt4,touvron2023llama,Bai2023QwenTR,DeepSeekAI2025DeepSeekR1IR,Sun2024HunyuanLargeAO}, have led to the development of numerous excellent vision-language models~\cite{li2023blip2,instructblip,liu2023llava,Chen2023InternVS,Bai2023QwenVLAV}. These models predominantly target tasks involving static images and text, such as image-text dialogue and text-to-image generation. Inspired by these achievements, LVLMs for video understanding~\cite{li2023llamavid,maaz2023videochatgpt,lin2023videollava,zhang2023videollama,li2023videochat,ye2023mplug,li2023otter} have emerged, extending the scope of multi-modal tasks to dynamic scenes. To optimize these models effectively, QA pairs based on video clips are required. Consequently, large volumes of videos from all sources have been utilized to construct training datasets.

Despite the abundance and diversity of video data available on various online platforms, most of them are unstructured and noisy, making unified collection a significant challenge. 

Furthermore, substantial human efforts and time are also required for the latter high-quality annotation and filtering process.
Additionally, current LVLMs for video understanding are typically trained on large-scale, pre-constructed datasets~\cite{chen2011msvd,xu2016msrvtt,caba2015activitynet,castro-etal-2020-lifeqa}, targeting general-purpose scenarios.

However, in the practical deployment, LVLMs follow a ``pre-training\&fine-tuning'' manner to apply for specific downstream domains, such as movie understanding, security surveillance, or behavior recognition, where collecting and constructing tailored datasets remains a formidable challenge.

To address these challenges, we consider generating customized video datasets on demand. Inspired by the advances of generation models such as diffusion models, we explore the automatic generation of comprehensive datasets for video instruction tuning. Specifically, we propose \textbf{DreamFrame}, a novel framework that automatically synthesizes style-consistent video keyframes and corresponding QA pairs. Our framework comprises three primary stages: \\
1)\textbf { Movie Plot Generation.}
An LLM is first utilized to generate structured movie plots including movie prior information, frame descriptions and plot-related QA pairs. To mitigate context length limitations of LLMs, a story expansion strategy is proposed.
\\
2)\textbf { Style Immobilization Process.}
Then, to ensure visual consistency across generated frames, a style embedding is learned based on a diffusion model. This approach later guides the diffusion model to generate keyframes in a consistent style.\\
3)\textbf { Video instruction data generation.}
By integrating the previously generated frame descriptions with the learned style embeddings, we generate style-consistent key frames, resulting in a comprehensive instruction tuning corpus.

\begin{figure}[t]
\centering
\includegraphics[width=\columnwidth]{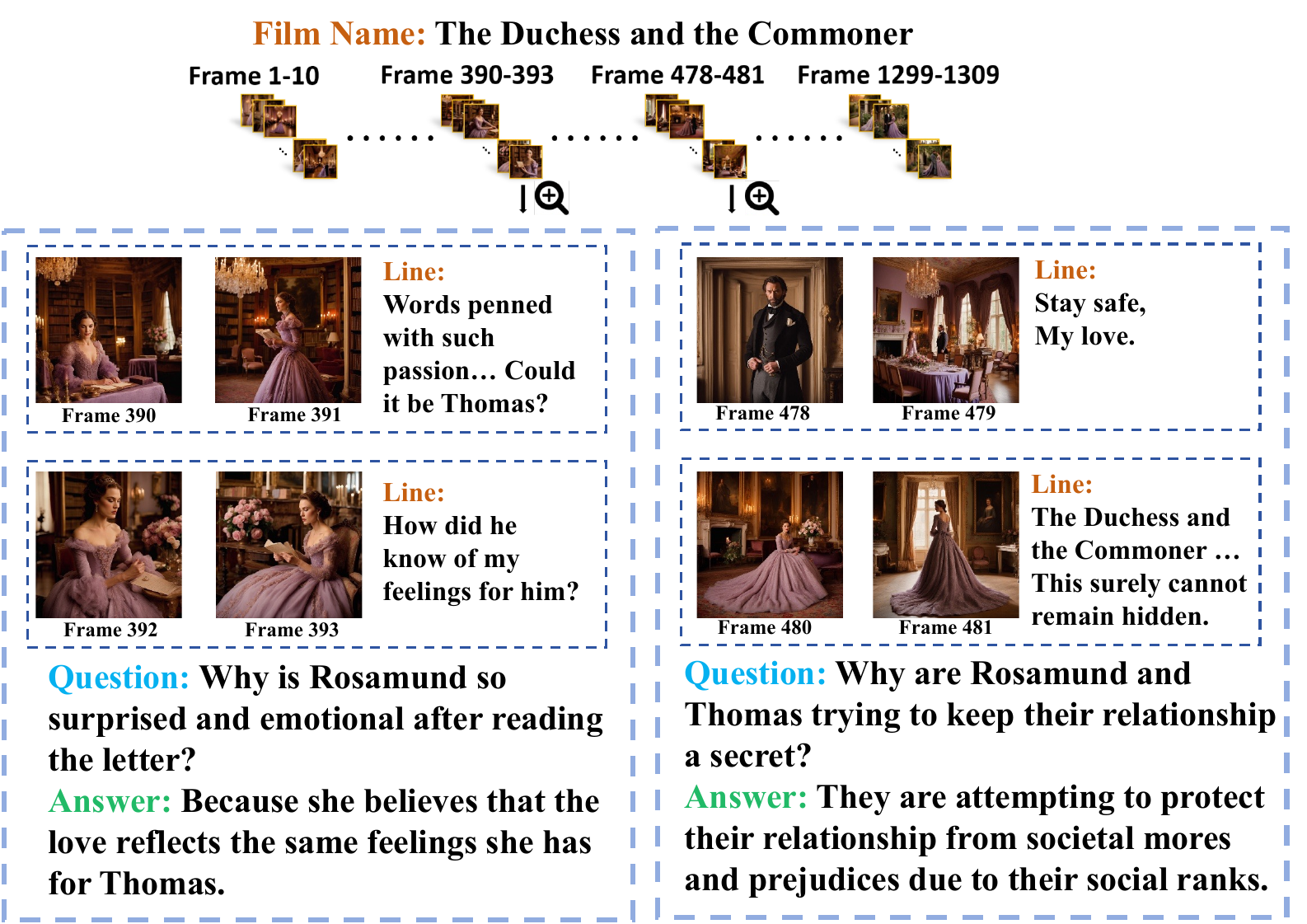}

\caption{Examples of generated data. DreamFrame generates consistent key frames with reasonable lines and corresponding question-answer pairs. These data are used to fine-tune multi-modal large language models for video understanding.}

\label{fig:teaser}

\end{figure}

Our methodology provides an innovative solution for the limitations of current datasets: 1) DreamFrame allows for the generation of datasets without constraints on data volume, ensuring a high degree of diversity within the generated content. 2) DreamFrame facilitates automatic annotation, significantly reducing the need for manual labor and associated costs. These advantages enhance the scalability, richness, and efficiency of dataset creation for video understanding. We hope that our preliminary exploration can offer promising insights for future research on automatic construction of video understanding datasets. Examples from our generated data are shown in~\autoref{fig:teaser}.

Our contributions are summarized as follows:
\begin{itemize}
\item We propose a novel framework, i.e., DreamFrame, for generating video instruction tuning datasets. A story expanding strategy and a style immobilized strategy are designed to ensure the quality of the generated data.
\item Leveraging our generative approach, we have developed and will publicly release a comprehensive dataset for video understanding, alongside a superb LVLM, i.e., DreamFrame-7B, trained for enhanced video understanding.
\item Extensive experiments are conducted to demonstrate that DreamFrame effectively and consistently enhances the capability of diverse LVLMs in video understanding. Meanwhile, our DreamFrame-7B remarkably outperforms previous LVLMs across different benchmarks, providing a new high-performance model for video understanding.

\end{itemize}


\section{Related Work}
\label{sec:rel}

\textbf{Vision Language Models.} With the achievements of large language models (LLMs) such as GPT-4 \cite{gpt4} along with their open-source alternatives like LLaMA \cite{touvron2023llama}, researchers focus on leveraging the advanced language abilities of LLMs and developing the vision language models (VLMs) that integrate vision models with LLMs for cross-modality understanding. Representative VLMs like LLaVA \cite{liu2023llava}, Flamingo \cite{Alayrac2022FlamingoAV} and InstructBLIP \cite{instructblip} have shown great capabilities in visual chat by constructing high-quality image-instruction pairs to align the image and text dimensions. Further, VLMs are developed for video understanding \cite{jin2024videolavit,li2023videochat,XVLM,VideoGraph}. Video-LLaMA \cite{zhang2023videollama} utilizes BLIP-2 \cite{li2023blip2} fuse video embedding using Q-Former. Video-ChatGPT \cite{maaz2023videochatgpt} extracts video embedding by averaging frame-level features across temporal and spatial dimensions respectively. LLaMA-VID~\cite{li2023llamavid} is proposed for video understanding by encoding each frame with only two tokens. Recent LVLMs like QwenVL series comes with superior video comprehension ability. However, most of these LVLMs are trained on general-purpose tasks, with little or no fine-tuning for specific domains. A potential cause of this limitation is the lack of domain-specific datasets.

\textbf{Video Instruction Tuning Datasets.} Current training datasets for video understanding are predominantly constructed by annotating samples from existing datasets~\cite{caba2015activitynet,chen2011msvd,xu2016msrvtt,YouCook2,castro-etal-2020-lifeqa,castro-etal-2022-wild}, with annotations typically generated through manual labeling or semi-automated methods. In contrast, our work is the first to explore fully automated dataset generation for video understanding, a direction that remains largely unexplored in the current literature.

\section{Method}
\label{sec:met}

\begin{figure*}[t]
\centering
\includegraphics[width=\textwidth]{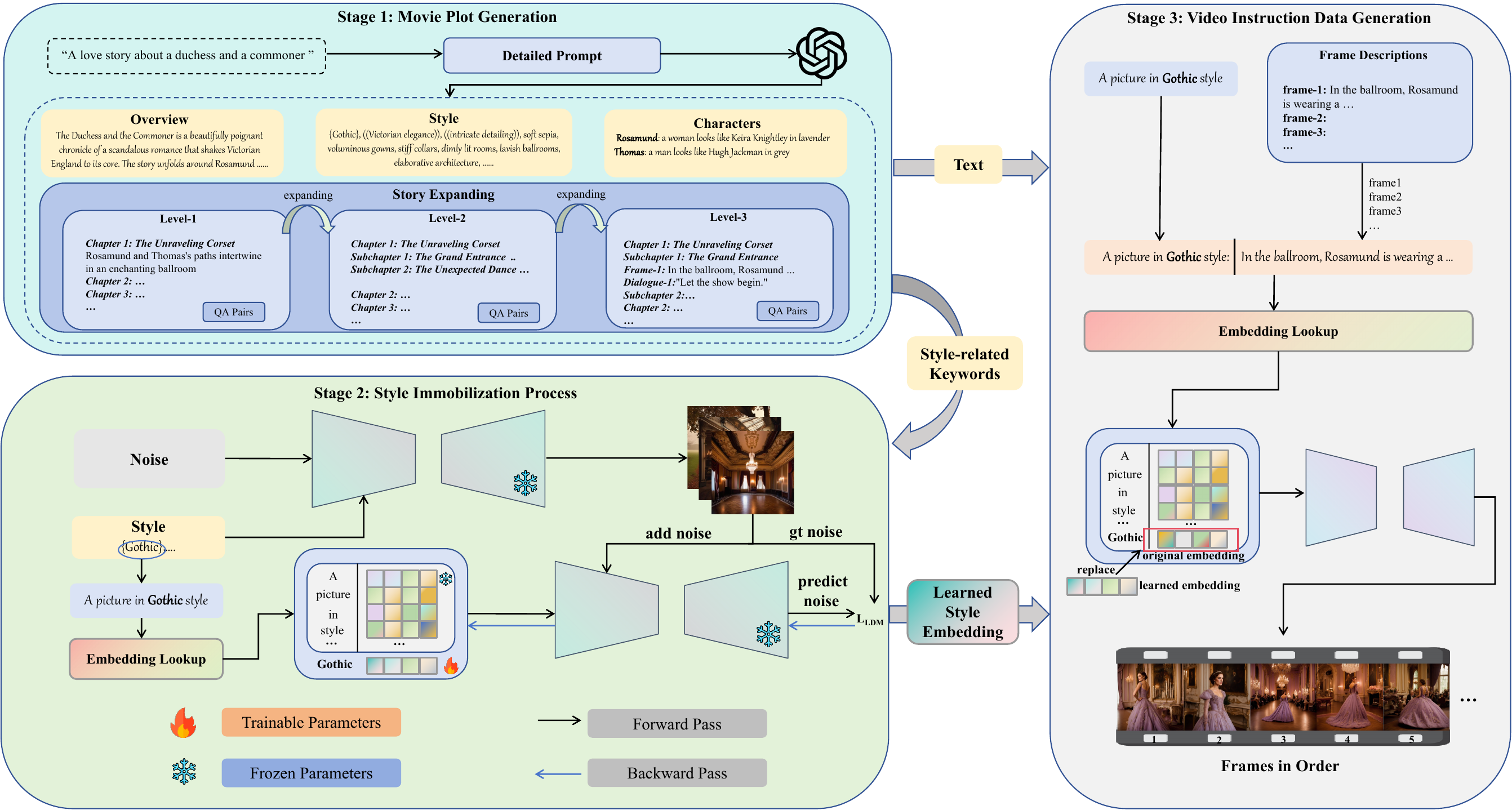} 
\caption{Our proposed pipeline for generating video instruction tuning datasets. With merely a simple thematic description, our pipeline is capable of generating key frames of an entire film. The pipeline can be roughly divided into three stages: (a) movie plot generation, where we generate the whole movie plot based on a theme phrase. (b) style immobilization process, where we learn a style embedding to immobilize the style-related keywords generated from the plot into the latent space of the diffusion model, guiding it to generate frames with fixed style. (c) video instruction data generation, where we integrate all the previously obtained information to ultimately generate consistent key frames.
}
\label{fig_method}
\end{figure*}

The overall framework of the proposed DreamFrame is shown in~\autoref{fig_method}, which consists of three stages: Movie Plot Generation, Style Immobilization Process and Video Instruction Data Generation.
\subsection{Movie Plot Generation} 
The primary objective of this step is to generate diverse and compelling movie plots. Each movie plot comprises elements including overview, movie style, characters and frame descriptions, which also serves for the latter process. Here, to generate suitable textual plots, we enable GPT to adapt to our task through few-shot in-context learning with detailed prompts. Our ultimate goal is to generate consistent key frames. To achieve this, a detailed description is needed for each frame. Instinctively, all aforementioned generated textual information can be input into LLM $G$ with an instruction $I$ to gain the results at once. However, the consistency between each frame description can not be assured because of the long memory length limitation problem of LLMs. Generating all key frame descriptions simultaneously often results in insufficient context, leading to inconsistent and poor-quality text. To address this issue, we introduce a strategy using LLM to progressively expand the overview of the story into detailed frame descriptions. This expansion process is divided into three levels. Given the overview $O$ and instruction $I$, LLM $G$ is used to predict level-1 descriptions:
\begin{equation}
    C_1,C_2,...,C_i,...,C_l,Q_{L1} = G\left ( O,I  \right )
\end{equation}
where $C_i$ denotes the current level-1 description, $Q_{L1}$ denotes the corresponding QA pairs for level-1 and $l$ denotes the required length of the descriptions. 

Based on $C_i$, level-2 descriptions are then further predicted:
\begin{equation}
    \resizebox{0.7 \hsize}{!}{$S_{i,1},S_{i,2},...,S_{i,j},...,S_{i,m},Q_{L2} = G\left ( C_i,O,I  \right )$}
\end{equation}
where $S_{i,j}$ denotes the current level-2 description, $Q_{L2}$ denotes the corresponding QA pairs for level-2 and $m$ denotes the required length of the descriptions.

Finally, based on $C_i$ and $S_{i,j}$, desired frame descriptions are predicted:
\begin{equation}
    \resizebox{0.89 \hsize}{!}{$F_{i,j,1},F_{i,j,2},...,F_{i,j,k},...,F_{i,j,n},Q_{L3},D= G\left (S_{i,j} ,C_i,O,I  \right )$}
\end{equation}
where $F_{i,j,k}$ denotes the current frame description, $Q_{L3}$ denotes the corresponding QA pairs for level-3, $D$ denotes the corresponding dialogues and $n$ denotes the required length of the frame descriptions.

We then get totally $l \times m \times n$ frame descriptions and $len(Q_{L1})+len(Q_{L2})+len(Q_{L3})$ QA pairs.

\subsection{Style Immobilization Process}

Now the issue lays in how to convert text information from the first stage into corresponding visuals. Simply employing a T2I model like stable diffusion for image generation does not guarantee consistency between visual frames. The key to maintaining consistency lies in ensuring stylistic uniformity throughout the movie. To achieve this, techniques like DreamBooth \cite{Ruiz2022DreamBoothFT} are generally introduced. However, these methods usually need fine-tuning the entire model, which is clearly impractical when generating movies in bulk. Therefore, in light of these issues, we propose a method based on textual inversion to satisfy the requirements of both consistency and efficiency.

Textual inversion is typically employed in conjunction with Latent Diffusion Models (LDMs). As is known, the goal of LDMs is to remove the noise added to a latent representation of
an image. The LDM loss is given by: 
\begin{equation}
\resizebox{0.89 \hsize}{!}{$
    L_{LDM} = \mathbb{E}_{z\sim\mathcal{E}(x), y, \epsilon \sim \mathcal{N}(0, 1), t }\Big[ \Vert \epsilon - \epsilon_\theta(z_{t},t, c_\theta(y)) \Vert_{2}^{2}\Big] \,$}
    \label{eq:LDM_loss}
\end{equation}
where $t$ is the time step, $z_{t}$ is the latent noise, $\epsilon$ is the unscaled noise sample, $\epsilon_\theta$ is the denoising network, $c_\theta$ is the encoder for conditioning, $y$ is the conditioning input.

Assume $c_\theta$ is a text encoder, text input $y$ is first converted to tokens, which are indexes in some pre-defined dictionary. Each token is then linked to a unique embedding vector that can be obtained through an index-based lookup. These embedding vectors are typically learned as part of the text encoder $c_\theta$. The embedding space is the optimized objective of textual inversion. Now, in the first stage of DreamFrame, we generate a word that describes the style of the movie like ``\textbf{Gothic}''. We use this word to represent the style concept we wish to learn. The learned embedding vector $V_*$ will be used to replace the associated vector of the word like ``\textbf{Gothic}''. Then we can include the word that describes the style in a sentence like ``Generate an image in \textbf{Gothic} style: ...'' which follows by the actual content of the frame such as ``In the ballroom, Rosamund is talking...''. In this way, the specific style word ``\textbf{Gothic}'' will ``trigger'' the learned style embedding as conditioning input to ensure that the content of the frames adhere to a consistent style while still allowing for content variation. Then by assigning a style-related keyword to each film and learning a style embedding associated with that keyword, the style consistency of the movie frames can be largely ensured. 

To obtain the learned embedding as mentioned, a set of images (we use 10) which depicts the target style are needed. We generate those images using the movie style from stage one. The embedding related to the style is then learned through direct optimization, by minimizing the LDM loss of Equation \eqref{eq:LDM_loss} over images from the set. The optimization goal can be defined as:
\begin{equation}
  \resizebox{0.89 \hsize}{!}{$ v_* = \argmin_v \mathbb{E}_{z\sim\mathcal{E}(x), y, \epsilon \sim \mathcal{N}(0, 1), t }\Big[ \Vert \epsilon - \epsilon_\theta(z_{t},t, c_\theta(y)) \Vert_{2}^{2}\Big]$}
    \label{eq:v_opt}
\end{equation}
while $c_\theta$ and $\epsilon_\theta$ is fixed. This is a reconstruction task that motivate the learned embedding to capture fine visual details unique to the style concept.

Now, by our method, every movie can have their own style-related embedding through just minutes of training.

\subsection{Video Instruction Data Generation}
Based on previous two stages, we utilize the style embeddings to guide stable diffusion in generating key frames according to the key frame descriptions. More specifically, to generate key frames that are consistent in both characters and scenes, we initially replace character names in the frame description with corresponding celebrities (chosen by GPT-4). Following that, a style-immobilized embedding linked to a special token is utilized. This style embedding can serve as a condition to guide the stable diffusion model in generating scenes in a fixed style. This process is triggered by a special token, which denotes a specific style, such as ``\textbf{Gothic}''. Hence, by incorporating a sentence like ``generate an image in \textbf{Gothic} style:'' at the beginning of our frame description, combined with character names, we can generate style-consistent key frames. We construct around 1K videos. Detailed distribution of our dataset, please refer to our supplementary material.

\section{Experiment}
\label{sec:exp}
\subsection{Setup}
\label{subsec:setup}

\begin{table}[t]
    \centering
  
    \caption{\textbf{The gain from tuning with our dataset is universal among model architectures.} The best results are \textbf{bold}.}
    \label{tab:improve}
    \resizebox{1.0\columnwidth}{!}{

        \begin{tabular}{l|ccc|c}
        \toprule
        Model              & MvBench    & VideoBench       & TempCompass   & Avg.          \\ \midrule
        VideoChatGPT-7B \cite{lin2023videollava}      & 32.7          & 38.5          & 42.4          & 37.9          \\
        VideoChatGPT-7B+Ours & \textbf{35.8} & \textbf{41.6} & \textbf{45.7} & \textbf{41.0} \\ \midrule
        VideoLLaVA-7B \cite{lin2023videollava}      & 42.9          & 34.5          & 49.9          & 42.4          \\
        VideoLLaVA-7B+Ours & \textbf{46.1} & \textbf{37.9} & \textbf{52.8} & \textbf{45.6} \\ \midrule
        LLaMA-VID-7B \cite{li2023llamavid}       & 41.5          & 36.5          & 44.2          & 40.7          \\
        LLaMA-VID-7B+Ours  & \textbf{45.1} & \textbf{40.1} & \textbf{48.1} & \textbf{44.4} \\ \midrule
        \end{tabular}
    }
\end{table}

\textbf{Datasets and Benchmarks.}
To thoroughly investigate how our high-quality video-caption dataset enhances the capabilities of LVLMs, we conduct comprehensive evaluations of the model across three multi-modal video benchmarks. MVBench~\cite{Li2023MVBenchAC} is designed to evaluate LVLMs on video tasks that go beyond single-frame understanding, comprising 4,000 QA pairs sourced from 11 public video datasets, such as TVQA and FunQA. VideoBench~\cite{Ning2023VideoBenchAC} aggregates a collection of approximately 15,000 QA pairs covering 10 evaluation aspects from 13 video question-answering datasets. TempCompass~\cite{tempcompass} evaluates the fine-grained temporal reasoning abilities of LVLMs, focusing on aspects like speed, direction, and attribute changes. It features 410 videos and 7,540 carefully curated instructions to highlight temporal understanding and interactive capabilities.

\begin{table*}[th]
    \centering
    \caption{Comparison with previous leading methods on MVBench. * denotes our evaluation results with the public
checkpoints. The best results are bold and the second-best results are underlined.}
    \resizebox{\textwidth}{!}{
        \begin{tabular}{l|cccccccccccccccccccc|c}
        \toprule
        Model   & AS & AP & AA & FA & UA& OE & OI & OS & MD & AL & ST & AC & MC & MA & SC & FP & CO & EN & ER & CI & Avg\\
        
        \midrule
        Otter-V-7B~\cite{li2023otter}   & 23.0 & 23.0 & 27.5 & 27.0 & 29.5 & 53.0 & 28.0 & 33.0 & 24.5 & 23.5 & 27.5 & 26.0 & {\ul 28.5} & 18.0 & 38.5 & 22.0 & 22.0 & 23.5 & 19.0 & 19.5& \cellcolor{gray!20}{26.8} \\
        
        mPLUG-Owl-V-7B~\cite{ye2023mplug}  & 22.0 & 28.0 & 34.0 & 29.0 & 29.0 & 40.5 & 27.0 & 31.5 & {\ul 27.0} & 23.0 & 29.0 & 31.5 & 27.0 & 40.0 & 44.0 & 24.0 & 31.0 & 26.0 & 20.5 & 29.5  & \cellcolor{gray!20}{29.7}\\

        LLaMA-Adapter~\cite{Zhang2023LLaMAAdapterEF}   & 23.0 & 28.0 & 51.0 & 30.0 & 33.0 & 53.5 & 32.5 & 33.5 & 25.5 & 21.5 & 30.5 & 29.0 & 22.5 & 41.5 & 39.5 & 25.0 & 31.5 & 22.5 & 28.0 & 32.0 & \cellcolor{gray!20}{31.7}\\
        
        VideoChatGPT-7B~\cite{maaz2023videochatgpt}   & 23.5 & 26.0 & 62.0 & 22.5 & 26.5 & 54.0 & 28.0 & {\ul 40.0} & 23.0 & 20.0 & 31.0 & 30.5 & 25.5 & 39.5 & \textbf{48.5} & 29.0 & 33.0 & 29.5 & 26.0 & 35.5 & \cellcolor{gray!20}{32.7}\\
        
        VideoLLaMA-7B~\cite{zhang2023videollama}   & 27.5 & 25.5 & 51.0 & 29.0 & 39.0 & 48.0 & 40.5 & 38.0 & 22.5 & 22.5 & 43.0 & 34.0 & 22.5 & 32.5 & 45.5 & {\ul 32.5} & 40.0 & 30.0 & 21.0 & {\ul 37.0} & \cellcolor{gray!20}{34.1}\\
        
        VideoChat-7B~\cite{li2023videochat}  & 33.5 & 26.5 & 56.0 & 33.5 & 40.5 & 53.0 & 40.5 & 30.0 & 25.5 & 27.0 & 48.5 & 35.0 & 20.5 & 42.5 & {\ul 46.0} & 26.5 & 41.0 & 23.5 & 23.5 & 36.0 & \cellcolor{gray!20}{35.5}\\

        LLaMA-VID-7B* \cite{li2023llamavid}  & 42.0 & {\ul 43.0} & {\ul 63.5} & 35.5 & {\ul 56.5} & {\ul 56.0} & 37.5 & 34.0 & 19.0 & 26.5 & {\ul 84.5} & 42.0 & {\ul 28.5} & 44.5 & 40.5 & 22.0 & 39.0 & {\ul 36.5} & {\ul 44.0} & 34.5 & \cellcolor{gray!20}{41.5}\\
        
        VideoLLaVA-7B* \cite{lin2023videollava}  & {\ul 44.5} & 42.5 & 58.0 & {\ul 38.5} & 52.5 & 54.0 & \textbf{47.5} & \textbf{41.0} & \textbf{29.0} & \textbf{31.5} & 82.0 & {\ul 45.0} & 26.5 & \textbf{53.0} & 41.5 & \textbf{33.0} & {\ul 41.5} & 27.5 & 37.5 & 31.5 & \cellcolor{gray!20}{\ul 42.9}\\

        \midrule
        DreamFrame-7B  &\textbf{48.0} & \textbf{46.0}   & \textbf{67.0} & \textbf{39.5} & \textbf{59.0}  & \textbf{58.5} & {\ul 41.5} & 36.5          & 25.5 & {\ul 30.5} & \textbf{ 87.5}    & \textbf{45.5}          & \textbf{30.5} & {\ul 48.5} & 44.5 & 26.0          & \textbf{42.5} & \textbf{40.0}          & \textbf{ 46.5}     &\textbf{37.5} & 
        \cellcolor{gray!20}{\textbf{45.1}}\\

        \bottomrule
      
        \end{tabular}
    }
    \label{tab:mvbench}
\end{table*}

\renewcommand{\arraystretch}{1.1}
\begin{table*}[!t]
    \centering

    \caption {\textbf{Comparison with previous leading methods on VideoBench.} * denotes our evaluation results with the public checkpoints. The best results are \textbf{bold} and the second-best results are \underline{underlined}.}

    \label{tab:vbench}
    \scalebox{1.0}{
        \begin{tabular}{l|ccccccc|ccc|ccc|c}
        \toprule
        \multicolumn{1}{l|}{\multirow{2}{*}{Model}} 
        & \multicolumn{7}{c|}{VEU}    
        &\multicolumn{3}{c|}{PKQA}
        &\multicolumn{3}{c|}{CDM}&\multirow{2}{*}{Avg}\\ \cmidrule{2-14}

        & ANet          & MSVD          & MSRVTT        & TGIF          & YC2      & UCF           & MOT           & TV            & MV            & NBA           & LE            & DM            & SQA3D                  \\ \midrule

         Video-LLaMA-7B \cite{zhang2023videollama}    & 39.9          & 41.2          & 34.1          & 31.3          & 28.9          & 27.6          & 16.7          & 24.8          & 32.4          & 26.2          & \textbf{60.6} & 49.1          & 31.2          & \cellcolor{gray!20}{32.8}          \\
        mPLUG-Owl-7B \cite{ye2023mplug}     & 41.5          & 42.5          & 36.3          & 31.7          & 27.1          & 22.8          & \textbf{27.8} & 24.0          & 30.2          & 25.1          & 33.3          & 51.0          & 32.0          & \cellcolor{gray!20}{33.2}          \\

        Valley-7B \cite{luo2023valley}         & 38.1          & 32.0          & 28.0          & 31.4          & 29.1          & 20.3          & 11.1          & 23.7          & 32.6          &  31.3   & 41.7          & {\ul 56.5}    & 33.3          & \cellcolor{gray!20}{34.0}          \\

        VideoLLaVA-7B* \cite{lin2023videollava}    & 44.3          & 34.2          & 30.3          & {\ul39.4}          & 30.7          & 19.5          & {\ul 22.4}    & 27.1          & 33.4          & 25.5          & 33.5          & 50.6          & {\ul 39.1}    & \cellcolor{gray!20}{34.5}          \\
        ChatUniVi-7B \cite{jin2023chatunivi}     & {\ul 49.0}    & 48.6          & 41.7          & \textbf{ 41.3}    & 29.0          & 28.3          & 16.7          & 23.1          & 33.6          & 25.7          & 38.9          & 53.1          & 29.1          & \cellcolor{gray!20}{35.3}          \\

         VideoChat-7B \cite{li2023videochat}     & 44.6          & 42.2          & 37.4          & 33.7          & 27.7          & 22.4          & \textbf{27.8} & 26.2          & 34.1          & 28.6          & 39.9          & 55.4          & 31.4          & \cellcolor{gray!20}{35.4}          \\

        LLaMA-VID-7B* \cite{li2023llamavid}     & 45.2          & 44.5          & 39.1          & 29.1          & 29.3          & 27.9          & 11.1          & {\ul34.1} & 32.5          & 28.9          & 36.1          & 47.8          & 36.8          & \cellcolor{gray!20}{36.5}          \\
        Otter-7B \cite{li2023otter}          & 44.3          & {\ul55.0}          & \textbf{47.0} & 34.3          & 32.7          & 22.4          & 16.7          & 27.7          & \textbf{37.1} & {\ul34.3} & {\ul 52.8}    & 48.7          & 29.7          & \cellcolor{gray!20}{37.5}          \\

        PandaGPT-7B \cite{chen2024panda}       & 45.0          & 50.4          & 44.6          & 29.7          & 33.0          & {\ul 33.0}    & 16.7          & 27.9          & \textbf{37.1} & 31.1          & 41.7          & 56.0          & 30.8          & \cellcolor{gray!20}{37.5}          \\
        
        VideoChatGPT-7B \cite{maaz2023videochatgpt}   & 46.6          & \textbf{57.5} & {\ul 46.3}    & 35.6          & \textbf{34.8} & 24.1          & \textbf{27.8} & 28.8          & {\ul 36.5}    & 22.5          & 41.7          & \textbf{58.2} & 37.2          & \cellcolor{gray!20}{{\ul 38.5}}    \\

         \midrule
        
        DreamFrame-7B & \textbf{49.3} & 47.9    & 44.7          & 34.3 & {\ul 34.6}    & \textbf{33.8} & 14.5    & \textbf{38.3}    & 35.4          & \textbf{34.8}          & 43.4          & 51.5          & \textbf{40.6} & \cellcolor{gray!20}{\textbf{40.1}} \\ \bottomrule
        \end{tabular}
}
\end{table*}
\renewcommand{\arraystretch}{1.15}
\begin{table*}[!t]
    \centering
    \caption {\textbf{Comparison with previous leading methods on TempCompass.} The best results are \textbf{bold} and the second-best results are \underline{underlined}.}
    \label{tab:tempcompass}
    \resizebox{\textwidth}{!}{
\begin{tabular}{l|ccccc|ccccc|ccccc|ccccc|c}
\toprule
\multicolumn{1}{l|}{\raisebox{-2\totalheight}[0pt][0pt]{Model}} &\multicolumn{5}{c|}{Multi-Choice QA}                                          & \multicolumn{5}{c|}{Yes/No QA}                                                & \multicolumn{5}{c|}{Caption Matching}                                         & \multicolumn{5}{c|}{Caption Generation} &\multirow{2}{*}{Avg.}  \\ \cmidrule{2-21} 
\multicolumn{1}{c|}{} &AC            & DI            & SP            & EV            & AT            & AC            & DI            & SP            & EV            & AT            & AC            & DI            & SP            & EV            & AT            & AC            & DI            & SP            & EV            & AT            \\ \midrule
Valley-7B \cite{luo2023valley} &47.0          & 29.3          & 32.5          & 18.9          & 29.9          & 58.1          & {\ul 52.0}    & {\ul52.5}          & 50.3          & {\ul 52.9}    & 15.5          & 21.4    & 22.0          & 28.3          & 22.9          & 24.7          &  20.4    & 21.9    & 35.8          &  29.4 & \cellcolor{gray!20}{33.4}    \\
PandaGPT-13B \cite{chen2024panda} &35.5          & 27.8          & 29.3          & 31.8          & 30.9          & 53.0          & 49.6          & 50.8          & \textbf{ 53.7}    & 52.2          & 56.6          & 51.4          & 44.3          & {\ul55.0}          & 49.0          & 23.7          & 25.7          & {\ul26.0}          & 29.8          & 32.6 & \cellcolor{gray!20}{40.4}         \\
VideoLLaMA-13B \cite{zhang2023videollama} &54.1          & 24.5          & 28.1          & 32.8          & 28.5          & {\ul68.1}          & 46.0          & 48.8          & 51.8          & 50.9          & 73.1          & 47.4          & 47.1          & 52.0          & 48.3          & {\ul 54.3}    & 21.3          & 13.9          & {\ul 38.5}    & 33.9 & \cellcolor{gray!20}{43.3}         \\
VideoChatGPT-7B \cite{maaz2023videochatgpt} &47.0          & 31.6          & 28.4          & {\ul37.1}          & 30.9          & 52.5          & 50.0          & 49.5          & 51.0          & 50.0          & 64.6          & 48.6          & 47.8          & 49.3          & 48.6          & 40.9          & 28.4          & 24.5          & 31.8          & 33.9 & \cellcolor{gray!20}{42.4}         \\
mPLUG-Owl-7B \cite{ye2023mplug} &{\ul66.6}          & 29.3          & 32.2          & 34.8          & {\ul 35.4}          & 64.4          & 50.6          & {\ul 51.2}          & 51.3          & 52.0          & 56.9          & 45.3          & 46.4          & 49.3          & 49.0          & 46.5          & 28.2          & \textbf{30.4}          & 31.2          & {\ul36.5} & \cellcolor{gray!20}{44.5}         \\

VideoLLaVA-7B \cite{lin2023videollava} &\textbf{ 70.4}          & {\ul32.2}          & \textbf{38.2}          & \textbf{41.4}    & \textbf{39.9}          & \textbf{ 74.3}    & 51.8          & 50.3          & 49.2          & 51.1          & \textbf{88.2}    & \textbf{ 53.8}    & \textbf{61.9} & \textbf{ 57.0}    & \textbf{ 58.3}    & 50.8          & {\ul28.7}          & 23.2          & 38.2          & 33.6 & \cellcolor{gray!20}{\textbf{ 49.9}}         \\
LLaMA-VID-7B \cite{li2023llamavid} &58.6          & 29.9          & 29.3          & 30.5          & 26.0          & 63.0          & 48.8          & 49.2          & 48.4          & 52.7          & 72.7          & 45.6          & 52.2          & 49.0          & 49.0          & 53.0          & 28.0          & 21.9          & 35.5          & 35.9 & \cellcolor{gray!20}{44.2}         \\ \midrule
DreamFrame-7B &61.3  & \textbf{32.7 }   & {\ul33.5} & 34.1 & 30.2 & 66.3 & \textbf{52.9} & \textbf{55.3} & {\ul53.4} & \textbf{55.1} & {\ul75.6} & {\ul49.1} & {\ul 56.5}    & 52.8 & {\ul53.9} & \textbf{57.2} & \textbf{32.8} & \textbf{25.4} & \textbf{40.1} & \textbf{39.7} &\cellcolor{gray!20}{{\ul48.1}} \\ \bottomrule
\end{tabular}    
}
\end{table*}

\begin{figure*}[t]
\centering
\includegraphics[width=\textwidth]{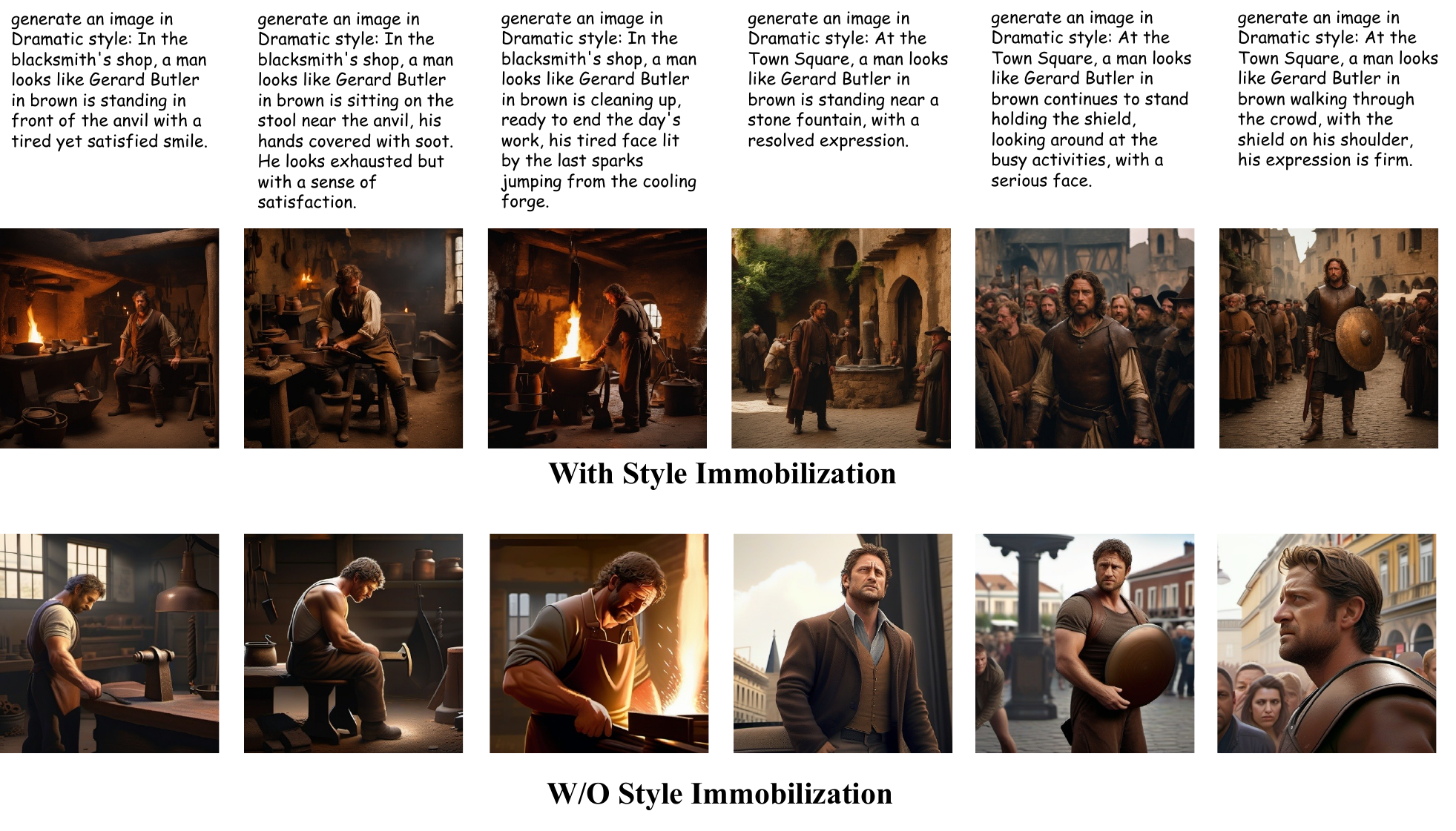} 
\vspace{-8mm}
\caption{Qualitative results of ablation study on style immobilization. The results indicate that the ablation of the strategy will cause obvious style inconsistency.
}
\label{fig_abla}
\end{figure*}

\textbf{Implementation Details.} We obtain our VLM model based on LLaMA-VID-7B~\cite{li2023llamavid} which provides basic abilities for video understanding (hereafter refered to as ``DreamFrame''). LLaMA-VID basically consists of a visual encoder, a text decoder, a projector and a LLM. For visual encoder, LLaMA-VID use EVA-G \cite{evaclip} as its ViT-based backbone and the patch size is set to 14. For text decoder, Qformer-7b \cite{instructblip} is used in our experiment. For projector, one-layer MLP is used to transform the embedding into context token. For LLM, we use vicuna-7b. During the model training phase, we employed the original LLaMA-VID configuration as the foundation for our training process. We utilized 2 NVIDIA A100 GPUs. To conserve GPU memory, we employed deepspeed with zero3 during model training, disabling tf32 and opting for fp16. The model under all settings is trained with a standard AdamW optimizer for 1 epoch. The initial learning rate is set as 2e-5 and updated by a cosine decay scheduler with a batch of 4 per GPU. Visual encoder and text decoder are both freezed during the training. Specifically, we conduct tuning after the second stage of LLaMA-VID. All video frames are input at 224 resolution. Please refer to our supplementary material for more details.

\subsection{Results on Video Understanding}

\begin{table}[]
\caption{Ablation study. We study the effects of proposed strategies in our method.}
\begin{center}{
 \resizebox{1.0\columnwidth}{!}{\begin{tabular}{cc|ccc}
\toprule
    Style Immobilization    &       Story Expanding  & \multicolumn{1}{l}{CLIP-Score $\uparrow$} & \multicolumn{1}{l}{LPIPS $\downarrow$}  \\ \midrule
 $\times$ & $\checkmark$ & 0.7175 & 0.7265 \\
 $\checkmark$ & $\times$ & 0.6902 & 0.5574 \\ 
  $\checkmark$ & $\checkmark$ & \textbf{0.8304} &\textbf{0.5316}     \\ 
 \bottomrule
\end{tabular}}
}
\end{center}

\label{table:abla}
\end{table}

\textbf{Enhancing Current LVLMs with DreamFrame.} We investigate the effectiveness of the video-QA data generated by DreamFrame to improve the performance of current
LVLMs. VideoChatGPT~\cite{maaz2023videochatgpt}, VideoLLaVA~\cite{lin2023videollava} and LLaMA-VID~\cite{li2023llamavid} are utilized as our base model. We further fine-tune their pre-trained models. For all three models, we only fine-tune the LLM component while keeping all other parts frozen. As shown in~\autoref{tab:improve}, our dataset consistently improves the alignment between video and language modalities in different LVLM architectures. Specifically, VideoLLaVA-7B achieves an average improvement of 3.1 across three comprehensive multimodal video benchmarks while VideoLLaVA-7B and LLaMA-VID-7B achieve an average improvement of 3.2 and 3.7.

\textbf{Comparison Results on Video Understanding.}  We compare our LVLM model DreamFrame-7B with other leading methods on three benchmarks as we mentioned. As shown in~\autoref{tab:mvbench}, DreamFrame outperforms the second-best method VideoLLaVA, by an average accuracy margin of 2.2\% on MVBench. Similarly,~\autoref{tab:vbench} reports the results on VideoBench, where DreamFrame achieves a 1.6\% improvement over the second-best method VideoChatGPT.
Results on TempCompass are presented in~\autoref{tab:tempcompass}. DreamFrame consistently performs favorably across most evaluation aspects, notably achieving the highest accuracy in all five sub-tasks under Caption Generation. Although DreamFrame ranks second in terms of overall average accuracy on TempCompass, it trails VideoLLaVA by only 1.8\%, indicating that the gap is minimal. Overall, these results demonstrate that DreamFrame obtains competitive performance across all three multi-modal video benchmarks, indicating its strong capacity for comprehensive video understanding.

\subsection{Ablation Study}

To verify the effectiveness of our proposed strategies in DreamFrame, we conduct an ablation study on two strategies: Story Expanding strategy, which is designed to ensure semantic consistency across frame descriptions, thereby preserving semantic consistency between generated frames and Style Immobilization strategy, which is aimed at maintaining visual style consistency between adjacent frames.
To evaluate these aspects, we use CLIP-Score~\cite{hessel-etal-2021-clipscore} to assess semantic consistency between image frames and LPIPS~\cite{lpips} to measure visual style consistency. As is shown in the second row of~~\autoref{table:abla}, the ablation of Story Expanding results in a significant drop of approximately 0.14 in CLIP-Score, while only causing a minor decrease of about 0.02 in LPIPS. This indicates that Story Expanding is a crucial component for ensuring semantic consistency across frames. The first row shows that removing Style Immobilization leads to a substantial performance degradation in LPIPS by around 0.17, highlighting its importance in maintaining visual style consistency. The qualitative results from~\autoref{fig_abla} also shows that the ablation of Style Immobilization leads to severe visual style inconsistency across frames.

\section{Conclusions}
In this paper, we propose an effective framework, DreamFrame, for generating style-consistent keyframes along with corresponding question–answer pairs, which can be used as tuning data for video understanding tasks. Extensive experiments validate the effectiveness of our approach. Beyond generating tuning dataset for video understanding, DreamFrame also holds potential for broader applications such as comic strip generation, artistic asset creation, and other domains that benefit from coherent multi-frame content generation. As an initial exploration of fully automatic training data generation, we hope that DreamFrame can inspire future research in this emerging field.

\newpage
\bibliographystyle{ACM-Reference-Format}
\bibliography{main}

\end{document}